\documentclass{article}
\usepackage{spconf,amsmath,epsfig}
\usepackage{float}
\usepackage{url}
\usepackage{multicol}
\usepackage{multirow}
\usepackage{adjustbox}
\usepackage[font=small,labelfont=bf]{caption}
\usepackage[table]{xcolor}

\newcommand\ourMethod{\text{FMARS}}

\definecolor{lightgray}{RGB}{240,240,240}

\title{\ourMethod: Annotating Remote Sensing Images for Disaster Management using Foundation Models}
\name{E. Arnaudo$^{1,2}$, J. L. Vaschetti$^{1,2}$, L. Innocenti$^{1}$, L. Barco$^{1,2}$, D. Lisi$^3$, V. Fissore$^3$, C. Rossi$^1$
    \thanks{This work was carried out in the context of the H2020 project SAFERS (GA n.869353), HEU project OVERWATCH (GA n.101082320) and Project NODES through the MUR—M4C2 1.5 of PNRR under Grant ECS00000036.}
}
\address{
    1. LINKS Foundation, \textit{AI, Data \& Space (ADS)}, Torino (TO), Italy \\
    2. Politecnico di Torino, \textit{Dipartimento di Automatica e Informatica  (DAUIN)}, Torino (TO), Italy \\
    3. Ithaca s.r.l., Torino (TO), Italy \\
}

\begin{document}
%
\maketitle
\begin{abstract}
Very-High Resolution (VHR) remote sensing imagery is increasingly accessible, but often lacks annotations for effective machine learning applications. Recent foundation models like GroundingDINO \cite{liu2023groundingdino} and Segment Anything (SAM) \cite{kirillov2023sam} provide opportunities to automatically generate annotations. This study introduces FMARS (Foundation Model Annotations in Remote Sensing), a methodology leveraging VHR imagery and foundation models for fast and robust annotation. We focus on disaster management and provide a large-scale dataset with labels obtained from pre-event imagery over 19 disaster events, derived from the Maxar Open Data initiative. We train segmentation models on the generated labels, using Unsupervised Domain Adaptation (UDA) techniques to increase transferability to real-world scenarios. Our results demonstrate the effectiveness of leveraging foundation models to automatically annotate remote sensing data at scale, enabling robust downstream models for critical applications. Code and dataset are available at \url{https://github.com/links-ads/igarss-fmars}.
\end{abstract}
\section{Introduction}
\label{sec:intro}
Remote Sensing (RS), and especially Very-High Resolution (VHR) images, represent a crucial resource for many real-world scenarios, including land use and land cover monitoring, urban planning, and disaster management.
Recent advances in satellite technologies have allowed for increasingly accessible remote sensing data, also thanks to public and private programs such as the European Union's Copernicus program \cite{copernicus} or the Maxar Open Data program \cite{maxaropendata}, which helps to democratize the access to medium and high-resolution data for research purposes on a global scale.
However, using such data with supervised machine learning mighe be challenging due to the limited availability of high-quality annotations.

In parallel, the machine learning landscape is witnessing the emergence of foundation models, which are large, general-purpose models that can adapt to different downstream tasks with minimal to no effort. This also involves computer vision, with models such as CLIP \cite{radford2021clip}, GroundingDINO \cite{liu2023groundingdino}, and Segment Anything \cite{kirillov2023sam}, which are able to provide robust image classification, object detection and segmentation capabilities in a wide range of contexts without further finetuning. Nevertheless, despite their remarkable performance, these solutions have often been employed in the context of natural images, with only a few attempts at applying them extensively on RS data at scale \cite{wang2023samrs}.

\begin{figure}[!t]
    \centering
    \includegraphics[width=\columnwidth]{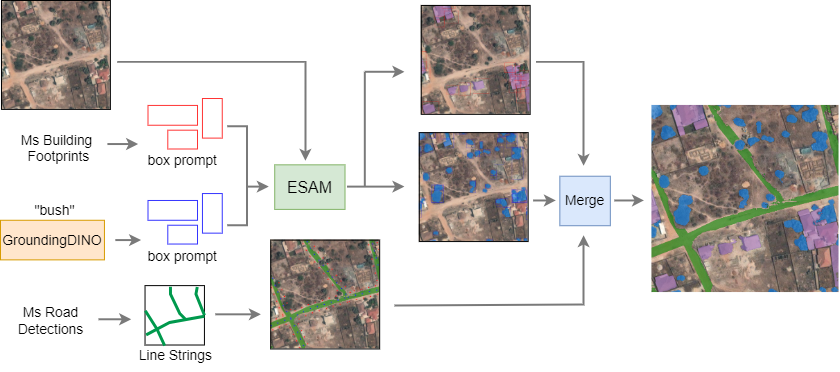}
    \caption{Annotation workflow adopted for the three selected classes. Each class is treated separately, with its own prompt construction pipeline, while the segmentation masks are extracted from the same image embeddings, and merged together in a single output.}
    \label{fig:workflow}
\end{figure}

In this work, we aim to bridge this gap between the growing availability of VHR remote sensing images and the potential of Vision Foundation Models (VFM) as robust annotators by designing an automated pipeline that combines open data sources with foundation models to generate either instance or semantic segmentation labels, starting from robust box prompts. Using this pipeline, named \ourMethod \ (i.e., Foundation Model Annotations in Remote Sensing), we automate the construction of a dataset designed for semantic labelling in damage assessment and disaster scenarios. The \ourMethod \ dataset includes 19 crisis events covered by Maxar Open Data imagery \cite{maxaropendata}, and comprises more than $25M$ annotations over a surface of over $125.000 \ km^2$. On this data, we provide instance-wise annotations subdivided into three example categories: buildings, roads, and high vegetation.
To validate the effectiveness of the annotation approach, we train state-of-the-art models on the generated labels, employing Unsupervised Domain Adaptation (UDA) techniques \cite{hoyer2022daformer,hoyer2023mic} for improved stability. Our results demonstrate the effectiveness of leveraging VFMs to automatically annotate remote sensing data at scale, enabling the development of smaller specific models for downstream applications.

\section{Related Work}

\begin{table*}[ht]
\begin{tabular}{|cccccccc|}
\hline
\textbf{Dataset}            & \textbf{\# Images}           & \textbf{Image size}  & \textbf{Resolution (cm)} & \textbf{Bands}          & \textbf{\# Instances}          & \textbf{\# Categories} & \textbf{Area ($km^2$)} \\ \hline
Vaihingen       & 33        & $~ 2,500 \times 2,500$    & 9             & IRRG & None & 6 & 1.33 \\
Potsdam         & 38        & $6,000 \times 6,000$      & 5 & RGBIR     & None & 6 & 11.08                       \\
iSAID           & 2,806     & ~ $4000 \times 4000$      & $\geq 50$     & RGB   & $655,451$   & 15    & 11,224    \\
xBD             & 9,168     & $1024 \times 1024$        & $\geq 50$     & RGB   & $> 700,000$ & 4   & 45,000        \\
SAMRS           & 105,090   & Mixed                     & $\geq 50$     & RGB   & $> 1.6M$    & Mixed   & Unknown   \\
\textbf{\ourMethod}   & 6,896    & $17,408 \times 17,408$                     & $\geq 30$     & RGB   & $> 25M$     & 3     & $> 125,000$ \\ \hline
\end{tabular}
\caption{Brief comparison between \ourMethod \ and similar VHR datasets available in literature.}
\label{tab:datasets}
\end{table*}

\label{sec:related}
Despite the recent advances in the computer vision field and the large data availability, remote sensing datasets remain limited in scope and scale compared to their natural images counterparts \cite{ghamisi2018potsdam-vai}, as shown in Table \ref{tab:datasets}. For instance, in downstream applications such as disaster management, the xBD dataset \cite{gupta2019xbd} focuses solely on building damage assessment, limiting its reuse in other contexts.
On the other hand, general-purpose datasets such as DOTA \cite{xia2018dota} may not be easy to adapt to particular downstream tasks, due to the limitations of the available annotations.
Considering models, Vision Foundation Models (VFM) have been successfully applied in several contexts, especially considering natural images \cite{caron2021dino,kirillov2023sam,liu2023groundingdino}, and downstream tasks in medical imagery \cite{cui2023medical1, mazurowski2023medical2}. Previous works have already assessed the applicability of VFMs to remote sensing images, including SAM \cite{ren2024sam-rs1}, applied to several semantic and instance segmentation tasks. Other attempts assessed the feasibility of using foundation models such as GroundingDINO \cite{liu2023groundingdino} for annotation purposes \cite{zhang2023text2seg}, or the combination of SAM with text prompts, encoded via CLIP \cite{radford2021clip}, to automatically generate segmentation masks for specific outputs \cite{osco2023sam-rs2}.
However, to the best of our knowledge, despite the large availability of remote sensing imagery, only a few attempts have been made to provide automated annotations at scale. SAMRS is the prime example \cite{wang2023samrs}, providing an extended set of annotations over well-known datasets such as DOTA \cite{xia2018dota} and DIOR \cite{zhan2023dior}.

\section{Materials and Methods}
\label{sec:method}


\begin{table}[!ht]
    \scriptsize
    \centering
    \begin{adjustbox}{width=1.0\columnwidth}
    \begin{tabular}{|l|l|r||l|l|r|}
    \hline
    \textbf{Event name} & \textbf{Year} & \textbf{Area ($km^2$)} & \textbf{Event name} & \textbf{Year} & \textbf{Area ($km^2$)} \\
    \hline
    Cyclone Mocha & 2023 & 3,446.4 & Morocco earthquake & 2023 & 49,901.9 \\
    Italy (Emilia) flooding & 2023 & 1,519.1 & Canada (NWT) wildfires & 2023 & 468.6 \\
    Gambia flooding & 2022 & 391.2 & Sudan flooding & 2022 & 249.3 \\
    Hurricane Fiona & 2022 & 1,341.8 & Afghanistan earthquake & 2022 & 4,180.6 \\
    Hurricane Ian & 2022 & 30,743.2 & Cyclone Emnati & 2022 & 8,506.0 \\
    Hurricane Idalia & 2023 & 12,156.4 & Kentucky flooding & 2022 & 1,641.6 \\
    India floods & 2023 & 496.3 & Pakistan flooding & 2022 & 7,528.7 \\
    Indonesia earthquake & 2022 & 1,011.3 & Georgia landslide & 2023 & 157.4 \\
    Turkey earthquake & 2023 & 2,745.7 & South Africa flooding & 2022 & 559.7 \\
    Kalehe flooding & 2022 & 89.9 & & & \\
    \hline
    \end{tabular}
    \end{adjustbox}
    \caption{List of events included in the \ourMethod \ dataset, including its year and total surface coverage derived from VHR imagery.}
    \label{tab:events}
\end{table}

\subsection{Foundation Models}
We adopt a combination of two large vision models for the annotation process, namely Segment Anything (SAM) \cite{kirillov2023sam}, in its resource efficient variant \cite{xiong2023efficientsam}, and GroundingDINO \cite{liu2023groundingdino}.
At its core, SAM and its derivatives are standard transformer-based segmentation networks that have been trained using \textit{promptable segmentation}. In contrast with other segmentation objectives, this task receives two inputs: an image and a prompt. While the former is processed using a large and robust image encoder, the latter is embedded into the decoder using a prompt encoder, and exploited as query by a lightweight mask decoder that produces segmentation masks. To resolve ambiguities, SAM can predict multiple outputs with its associated confidence for the same inputs.
The prompts can be extremely flexible, ranging from sparse inputs such as a single point, a bounding box, or text, to dense arrays such as a binary mask. While points and boxes are encoded as simple positional embeddings, text is processed using off-the-shelf models such as CLIP \cite{radford2021clip}, and masks are combined with the encoded image using a series of convolutions and element-wise sums.
Following previous works \cite{mazurowski2023medical2}, we adopt box prompts in our mask generation process, given its robustness and flexibility. This also combines naturally with the inputs at our disposal, comprising open data sources for buildings and roads (see Section \ref{sec:data-sources}), and box object detections derived from GroundingDINO.
This model introduces cross-modal fusion between a text prompt and an image to provide open-set object detection capabilities, using BERT as text processor \cite{devlin2019bert} and a Swin Transformer \cite{liu2021swin} as image encoder. 
While outputs may be approximate compared to human annotations, GroundingDINO provides a huge flexibility to generate bounding boxes for potentially any known object, given a text prompt. This allows us to obtain first estimates for objects not having a ground truth, such as vegetation, and thus exploit GroundingDINO as a prompt generator for the subsequent SAM masking phase \cite{osco2023sam-rs2}.

\subsection{Data Sources}
\label{sec:data-sources}

 Considering fine-grained segmentation in disaster management contexts, VHR imagery becomes necessary since lower-resolution satellite sources such as Copernicus Sentinel-2 do not provide enough image content to characterize objects of interest, such as buildings, or roads.
 To this date, the largest source of disaster-related VHR imagery is represented by the Maxar Open Data Program \cite{maxaropendata}. This initiative provides pre- and post-event RGB images from more than 100 major crisis events since 2017 worldwide, with a total surface coverage of more than $2.6M \ km^2$.
 We select a subset of resources containing RGB imagery, obtaining 19 events, spanning from 2022 to 2023, as displayed in Table \ref{tab:events}, and summing up to an area of $127,134 \ km^2$. 
 Inspired by current state-of-the-art disaster management datasets \cite{gupta2019xbd}, we focus our dataset construction process on infrastructures, namely buildings and roads, which are often the focus in post-event damage assessment, and high vegetation, which usually occludes the underlying surface.
 Among open resources providing infrastructure information, we select the Microsoft's Building Footprints and Road Detection datasets \footnote{https://github.com/microsoft/GlobalMLBuildingFootprints}, which contain building footprints polygons and road graphs on a global scale generated by applying deep learning models on VHR satellite imagery, respectively.
For buildings, we do not directly adopt them as ground truth labels, but rather we exploit them as trustworthy yet approximate prompts for the SAM model.
Lacking a reliable source to derive high vegetation prompt from, for such category we adopt GroundingDINO as our bounding box generator.

\subsection{Annotation Workflow}

We aim to generate segmentation labels for three classes: buildings, roads, and high vegetation.
While disaster risk management mainly focuses on damage assessment by comparing pre- and post-event images, we first concentrate our efforts on delineating infrastructures on pre-event acquisitions only. In fact, damage assessment frameworks typically delineate relevant entities in pre-event images, using the post-event image to determine the sustained damage \cite{gupta2019xbd,shen2021bdanet}. We argue that the first phase (i.e., identifying the exposed elements before the event) is crucial for any subsequent analysis, while the damage assessment could be carried out considering the output of the first phase and the post-event image using an ad-hoc model.
Considering buildings, we generate box prompts by simply extracting axis-aligned bounding boxes (AABB) from each footprint polygon. On the other hand, road graphs represent a challenge for prompt-based segmentation because their sparse lattice does not allow for fine contour generation. In this case, the point-based prompts did not yield satisfactory results, therefore we opted to simply rasterize the available vector lines with a predefined buffer radius of 5m.
For vegetation, we derive boxes using GroundingDINO with simple text queries like \textit{green trees} or \textit{bushes}, observing better performance on trees with the latter, likely due to the aerial viewpoint occluding the tree trunk, which is uncommon in natural images.
In order to ensure a certain degree of confidence for the generated outputs, we use a minimum box threshold of $0.12$ and a text threshold of $0.3$. We further filter out noisy outputs by applying non-maxima suppression (NMS) at 0.5, removing boxes with aspect ratio lower than 1:2, and maximum area over $7000 \ m^2$.
Similar to buildings, we then use the generated boxes as prompts for SAM to extract segmentation labels.
Last, we store the resulting delineation and its class as a single vector polygon to allow for both instance or semantic segmentation tasks.

\subsection{Experiments}
\begin{table}[!h]
   \centering
   \begin{adjustbox}{width=1.0\columnwidth}
\begin{tabular}{|l|cc|cc|cc|cc|c|c|}
\hline
\multirow{2}{*}{\textbf{Method}}                       & \multicolumn{2}{c|}{\textbf{Background}} & \multicolumn{2}{c|}{\textbf{Roads}} & \multicolumn{2}{c|}{\textbf{High Veg.}} & \multicolumn{2}{c|}{\textbf{Buildings}} & \multirow{2}{*}{\textbf{mAcc.}} & \multirow{2}{*}{\textbf{mIoU}} \\
                                                       & \textbf{Acc.}       & \textbf{IoU}       & \textbf{Acc.}     & \textbf{IoU}    & \textbf{Acc.}       & \textbf{IoU}      & \textbf{Acc.}       & \textbf{IoU}      &                                 &                                \\ \hline
       \rowcolor{lightgray} SegFormer (base) & 72.91               & 61.41              & 0.11              & 0.10            & 7.60                & 1.33              & 0.00                & 0.00              & 20.15                           & 15.71                          \\
MIC                                                    & 44.79               & 42.47              & 55.94             & 29.89           & 64.45               & 10.56             & 82.47               & 21.33             & 61.91                           & 26.06                          \\
DAFormer                                               & 53.06               & 50.14              & 55.44             & 31.79           & 64.61               & 16.80             & 79.91               & 17.29             & \textbf{63.26}                  & \textbf{33.07}                 \\ \hline
\end{tabular}

   \end{adjustbox}
   \caption{Performance comparison of DAFormer and MIC on the FMARS dataset across different classes.}
   \label{tab:performance_comparison}
\end{table}

\begin{table}[!h]
   \centering
   \begin{adjustbox}{width=1.0\columnwidth}
   \begin{tabular}{|l|cc|cc|cc|cc|c|c|}
\hline
\multirow{2}{*}{\textbf{Method}}                   & \multicolumn{2}{c|}{\textbf{Background}} & \multicolumn{2}{c|}{\textbf{Roads}} & \multicolumn{2}{c|}{\textbf{High Veg.}} & \multicolumn{2}{c|}{\textbf{Buildings}} & \multirow{2}{*}{\textbf{mAcc.}} & \multirow{2}{*}{\textbf{mIoU}} \\
                                                   & \textbf{Acc.}       & \textbf{IoU}       & \textbf{Acc.}     & \textbf{IoU}    & \textbf{Acc.}       & \textbf{IoU}      & \textbf{Acc.}       & \textbf{IoU}      &                                 &                                \\ \hline
\rowcolor{lightgray} FMARS labels & 71.34               & 41.16              & 68,72             & 47.03           & 69.37               & 58.54             & 59.47               & 54.14             & 67.23                           & 50.22                          \\
SegFormer (base)                                   & 97.40               & 27.90              & 0.06              & 0.06            & 8.24                & 7.68              & 0.00                & 0.00              & 26.44                           & 8.91                           \\
MIC                                                & 76.59               & 36.21              & 44.84             & 40.15           & 51.78               & 48.52             & 63.54               & 56.41             & 59.19                           & 45.32                          \\
DAFormer                                           & 70.56               & 38.02              & 65.97             & 54.77           & 56.57               & 52.64             & 69.10               & 60.20             & \textbf{65.55}                  & \textbf{51.41}                 \\ \hline
\end{tabular}
   \end{adjustbox}
   \caption{Performance comparison of DAFormer and MIC on a manually labelled Maxar partition across different classes.}
   \label{tab:gt_comparison}
\end{table}

\begin{figure*}[h]
    \centering
    \includegraphics[width=1.0\textwidth]{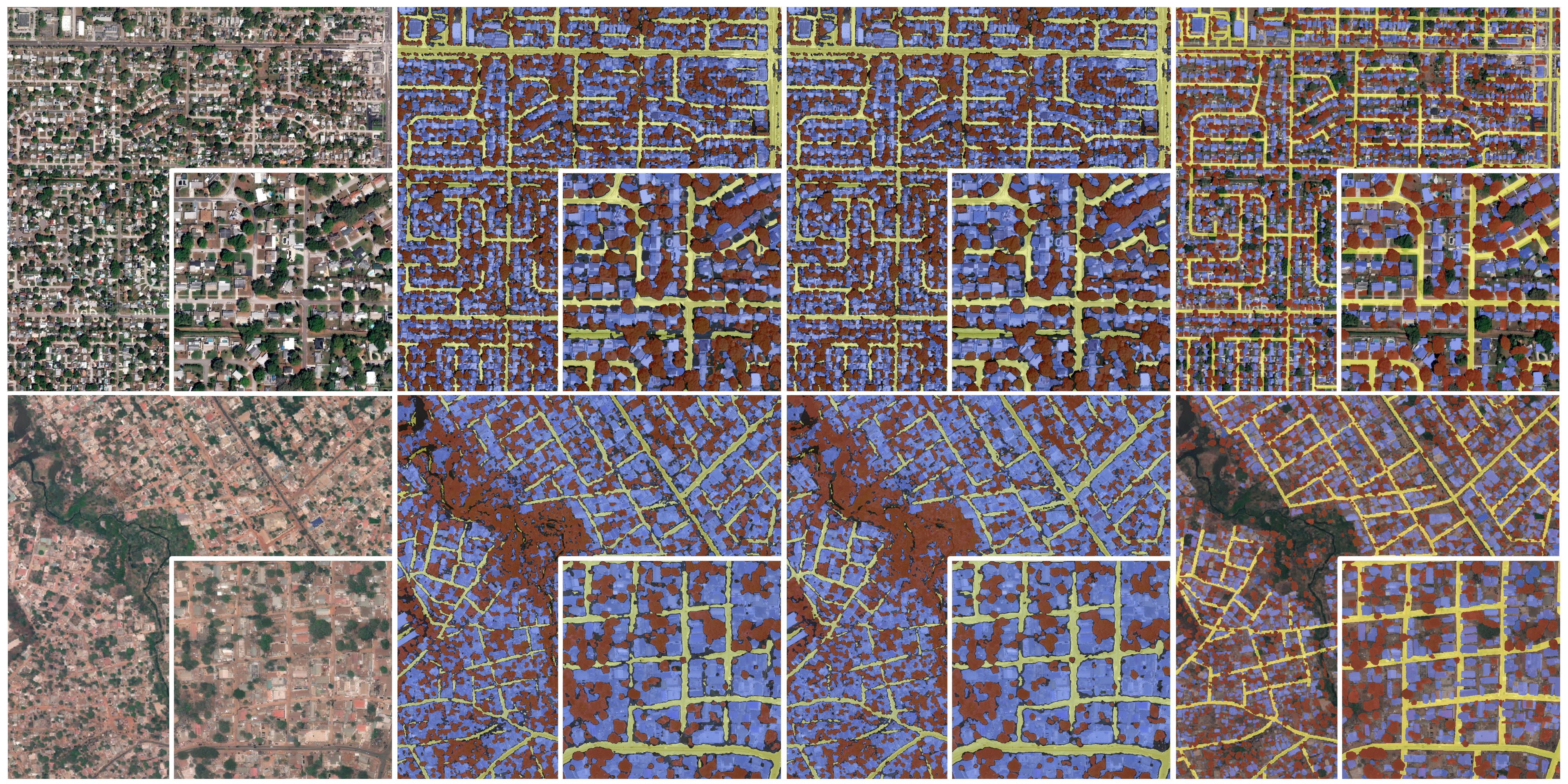}
    \caption{Qualitative results obtained over two example areas, namely USA (top) and Gambia (bottom). from left to right: RGB image, DAFormer, MIC, and \ourMethod \ ground truth. Best viewed zoomed in.}
    \label{fig:results}
\end{figure*}

Using the dataset annotated with our \ourMethod \ approach, we can train standard semantic segmentation models to evaluate the knowledge transfer ability to smaller and more deployable models.
In our experiments, we adopt state-of-the art solutions based on Segformer \cite{xie2021segformer}. To counteract the inherent inaccuracies in fully automated labeling and the consequently lower recall for categories such as high vegetation, we apply UDA techniques for improved stability during training. Specifically, we adopt DAFormer (\cite{hoyer2022daformer} and Masked Image Consistency (MIC) \cite{hoyer2023mic}, both based on self-training in a \textit{teacher-student} framework paradigm.
We select a separate full-size image from each event as our test set based on the average information content, for a total of 19 images, and we conduct a full training using pretrained ImageNet weights for the backbone components.
To address the high precision and lower recall of the generated labels, as well as missing categories, we train the models in an open set context, ignoring the background class \cite{oliveira2023fully}. As simple baseline, we apply a confidence threshold to the Softmax outputs, empirically evaluating the optimal cutoff threshold $\tau = 0.9$ for both models.
We conduct every experiment using a tile size of $512 \times 512$ and random sampling, weighted by the entropy (i.e., information content) of the available label, for $30,000$ iterations using AdamW as optimizer. For the UDA components, we maintain their original configuration, except for the removal of the ImageNet feature distance.

Given the automated pipeline and the low reliability of the obtained labels for performance measurement, we validate results against the FMARS labels, as well as a small sample of 45 manually labelled tiles, derived from crops of each image in the test set.
Tables \ref{tab:performance_comparison} and \ref{tab:gt_comparison} present the numerical results in terms of accuracy and Intersection over Union (IoU), evaluated against the FMARS test set and manual labels, respectively. The scores highlight the challenge of the problem at hand, where the baseline solution, without any precautions, collapses during training. Both UDA solutions appear very effective, with DAFormer even surpassing the original FMARS labels on the manually produced ground truth.
This is evident in the qualitative results shown in Fig. \ref{fig:results} that demonstrate a high fidelity between the model predictions and the ground truth, with DAFormer exhibiting more robustness to the challenging \textit{high vegetation} class. These findings highlight the effectiveness of \ourMethod \ as an automated technique to leverage foundation models for annotation tasks in remote sensing. With the necessary precautionary measures such as UDA techniques, the proposed pipeline allows the generation of accurate labels on a large scale, and even the knowledge transfer to smaller yet accurate downstream segmentation models in the absence of manually labeled datasets.

\section{Conclusions}
\label{sec:conclusions}

In this work, we propose \ourMethod, a pipeline for automated large-scale annotation of VHR imagery leveraging VFM. \ourMethod \ exploits the increasing availability of open-source images and the flexibility of promptable large models to automatically generate high-quality instance and semantic segmentation labels.
As an example, we focus on the critical domain of disaster management and construct the \ourMethod \ dataset, providing over 25 million annotations across 19 disaster events around the globe, derived from pre-event imagery from the Maxar Open Data initiative.
As representative downstream application, we train state-of-the-art segmentation frameworks on the generated labels. Due to the potential inaccuracies in the automated labels, we employ UDA techniques to increase the robustness of the learned features to real-world scenarios.
While the proposed approach demonstrates promising results, it has certain limitations. First, the dataset currently focuses only on pre-event images. Second, we limited the taxonomy to three classes in these tests. Finally, achieving high recall requires significant computational effort, especially in downstream training. Future work may address these limitations by developing a more robust automated annotation pipeline, or by exploring zero-shot learning approaches for open-set or panoptic segmentation tasks, with a potentially boundless taxonomy.



\vfill
\pagebreak

\bibliographystyle{IEEEbib}
{
\footnotesize
\bibliography{strings,refs}
}

\end{document}